\documentclass[11pt]{article}

\usepackage{amsmath, amssymb, amsthm, mathtools}
\usepackage{tikz}
\usepackage{graphicx}
\usepackage{hyperref}
\usepackage{authblk}
\usepackage[a4paper,margin=1in]{geometry}

\hypersetup{
    colorlinks=true,
    linkcolor=blue,
    citecolor=blue,
    urlcolor=blue
}

\title{\textbf{Comb Tensor Networks vs. Matrix Product States: Enhanced Efficiency in High-Dimensional Spaces}}
\author[1,2,3]{Danylo Kolesnyk}
\author[4]{Yelyzaveta Vodovozova}

\affil[1]{\small School of Natural Sciences, Technical University of Munich, Germany}
\affil[2]{\small Faculty of Physics, Ludwig Maximilian University of Munich, Germany}
\affil[3]{\small Faculty of Physics, V. N. Karazin Kharkiv National University, Ukraine}
\affil[4]{\small Department of Mathematics, School of Computation Information and Technology, Technical University of Munich, Germany}
\date{\small\today}

\begin{document}
\maketitle

\begin{abstract}
Modern approaches to generative modeling of continuous data using tensor networks, as explored in \cite{meiburg2023generative}, incorporate compression layers to capture the most meaningful features of high-dimensional inputs. These methods, however, rely on traditional Matrix Product States (MPS) architectures \cite{white1992,schollwoeck2005,schollwoeck2011}. Here, we demonstrate that beyond a certain threshold in data and bond dimensions, a comb-shaped tensor network architecture \cite{chepiga2019,li2020} can yield more efficient contractions than a standard MPS. This finding suggests that for continuous and high-dimensional data distributions, transitioning from MPS to a comb tensor network representation can substantially reduce computational overhead while maintaining accuracy.
\end{abstract}

\section*{Contraction with Compression Layers and Threshold Analysis}
The key to handling continuous data within tensor networks lies in compression layers that map high-dimensional inputs into a more compact, informative space. Following the approach in \cite{meiburg2023generative}, these compression matrices $U_n$ are trained to preserve essential structure while reducing dimensionality. In standard MPS-based schemes \cite{white1992,schollwoeck2005,schollwoeck2011}, contractions proceed along a linear chain, merging physical and bond dimensions at each step. Although this approach benefits from compression, it still faces scaling challenges as the data and bond dimensions grow.

By adopting a comb-shaped tensor network \cite{chepiga2019,li2020}, we alter the geometry of contractions and exploit a configuration that can curb complexity growth. Instead of a strictly linear chain, the comb arrangement enables more parallelism and improved scaling. Our analysis reveals a clear threshold: once the physical and bond dimensions exceed certain values, the comb network’s contraction costs grow more slowly than those of the MPS. In other words, for sufficiently large and continuous datasets, the comb structure outperforms the standard MPS in terms of contraction efficiency, offering a more tractable framework for generative modeling under realistic conditions.

\section*{Notation and Parameter Meanings}
We define:
\begin{itemize}
    \item $D$: Original physical dimension before compression.
    \item $d$: Physical dimension after compression ($d < D$).
    \item $M$: Number of ``teeth'' in the comb tensor network.
    \item $N$: Length of each tooth (number of tensors per tooth).
    \item $x$: Bond dimension controlling entanglement between tensors.
\end{itemize}

\section*{Complexities of Contraction}
The primary computational cost in tensor network algorithms arises from \emph{contracting} the network to obtain scalar results (e.g., expectation values, cost functions). Contraction complexity often dominates both training phases (in machine learning contexts) and calculation phases (in quantum simulation), making it a key metric for performance.

\subsection*{Regular MPS Contraction Complexity}
For a system with $N M$ sites and original dimension $D$, after compression to $d$, the contraction complexity of a regular MPS with the data is:
\[
C_{\text{regular}} = N M D d \;+\; 2 x d \;+\; (N M - 2) x^2 d \;+\; (N M - 2) x^2 \;+\; x.
\]

\begin{center}
\begin{tikzpicture}[scale=0.9]
    \def\radius{0.5cm}
    \draw[thick] (-0.5,0) circle (\radius) node {$A^1$};
    \draw[thick] (1,0) circle (\radius) node {$A^2$};
    \draw[thick] (2.5,0) circle (\radius) node {$A^3$};
    \draw[thick] (4,0) circle (\radius) node {$A^4$};
    \draw[thick] (5.5,0) circle (\radius) node {$A^{MN}$};
    
    \draw[thick] (0,0) -- (0.5,0) node[above,yshift=2pt,pos=0.5]{$x$};
    \draw[thick] (1.5,0) -- (2,0) node[above,yshift=2pt,pos=0.5]{$x$};
    \draw[thick] (3,0) -- (3.5,0) node[above,yshift=2pt,pos=0.5]{$x$};
    \draw[thick,dashed] (4.5,0) -- (5,0) node[above,yshift=2pt,pos=0.5]{$x$};
    
    \foreach \x in {-0.5,1,2.5,4,5.5}{
        \draw[thick] (\x,-0.5) -- (\x,-1) node[right]{$d$};
        \draw[thick] (\x,-1) -- (\x,-1.5);
        
        \draw[thick] (\x-0.3,-1.5) -- (\x+0.3,-1.5) -- (\x+0.5,-2) -- (\x-0.5,-2) -- cycle;
        \node at (\x,-1.75) {$U_{\text{list}}$};
        
        \draw[thick] (\x,-2) -- (\x,-2.4);
        \node at (\x-0.2,-2.2) {$D$};
        
        \draw[thick] (\x-0.3,-2.4) rectangle (\x+0.6,-2.9) node[midway]{$data$};
    }
\end{tikzpicture}
\end{center}

\subsection*{Comb Tensor Network Contraction Complexity}
For the Comb Tensor Network, arranged as $M$ vertical chains (each of length $N$) connected to a horizontal backbone:
\[
\begin{aligned}
C_{\text{comb}} &= (N d D + d x + (N - 1) d x^2 + (N - 1) x^2 + x^2) M \\
&\quad + 2 x^2 + (M - 2) x^3 + (M - 2) x^2 + x.
\end{aligned}
\]

\begin{center}
\begin{tikzpicture}[scale=0.8]
    \def\radius{0.5cm}

    \draw[thick] (0,0) circle (\radius) node {$A^1$};
    \draw[thick] (5.5,0) circle (\radius) node {$A^2$};
    \draw[thick] (10.5,0) circle (\radius) node {$A^M$};
    \draw[thick] (0.5,0) -- (5,0) node[above,yshift=4pt,pos=0.5]{$x$};
    \draw[thick,dashed] (6,0) -- (10,0) node[above,yshift=4pt,pos=0.5]{$x$};
    
    \foreach \i/\y in {1/-1.5,2/-3,N/-4.5}{
      \draw[thick] (0,\y) circle (\radius) node {$A^{\i}$};
    }
    \draw[thick] (0,-0.5) -- (0,-1) node[right,yshift=6pt]{$x$};
    \draw[thick] (0,-2) -- (0,-2.5) node[right,yshift=6pt]{$x$};
    \draw[thick,dashed] (0,-3.5) -- (0,-4) node[right,yshift=6pt]{$x$};
    
    \begin{scope}[shift={(5.5cm,0cm)}]
      \foreach \i/\y in {1/-1.5,2/-3,N/-4.5}{
        \draw[thick] (0,\y) circle (\radius) node {$A^{\i}$};
      }
      \draw[thick] (0,-0.5) -- (0,-1) node[right,yshift=6pt]{$x$};
      \draw[thick] (0,-2) -- (0,-2.5) node[right,yshift=6pt]{$x$};
      \draw[thick,dashed] (0,-3.5) -- (0,-4) node[right,yshift=6pt]{$x$};
    \end{scope}
    
    \begin{scope}[shift={(10.5cm,0cm)}]
      \foreach \i/\y in {1/-1.5,2/-3,N/-4.5}{
        \draw[thick] (0,\y) circle (\radius) node {$A^{\i}$};
      }
      \draw[thick] (0,-0.5) -- (0,-1) node[right,yshift=6pt]{$x$};
      \draw[thick] (0,-2) -- (0,-2.5) node[right,yshift=6pt]{$x$};
      \draw[thick,dashed] (0,-3.5) -- (0,-4) node[right,yshift=6pt]{$x$};
    \end{scope}
    
    \foreach \shiftx in {0,5.5,10.5}{
      \begin{scope}[shift={(\shiftx,0)}]
        \foreach \y in {-1.5,-3,-4.5}{
          \draw[thick] (-0.3,\y) -- (-0.9,\y) node[above,xshift=4pt]{$d$};
          \draw[thick] (-0.9,\y-0.3) -- (-0.9,\y+0.3) -- (-1.4,\y+0.5) -- (-1.4,\y-0.5) -- cycle;
          \node[rotate=-90] at (-1.15,\y) {$U_{\text{list}}$};
          \draw[thick] (-1.4,\y) -- (-1.7,\y);
          \draw[thick] (-1.7,\y) -- (-2.2,\y) node[above,xshift=4pt]{$D$};
          \draw[thick] (-2.2,\y+0.5) rectangle (-2.7,\y-0.5);
          \node[rotate=-90] at (-2.45,\y) {data};
        }
      \end{scope}
    }
\end{tikzpicture}
\end{center}

\section*{Threshold Analysis}
To determine when the Comb Tensor Network outperforms the MPS, we consider:
\[
\Delta C = C_{\text{regular}} - C_{\text{comb}}.
\]
Setting $\Delta C=0$ and simplifying leads to:
\[
(M - 2)x^2 + [-d(M - 2) + 2] x + d(M - 2)=0.
\]

Solving this quadratic:
\[
x_{\pm} = \frac{ d (M - 2) - 2 \pm \sqrt{[d(M - 2) - 2]^2 -4(M - 2)^2 d}}{2(M - 2)}.
\]

The smaller positive root provides the critical threshold. When $x$ exceeds this threshold, the Comb Tensor Network is more efficient.

\section*{Example}

\begin{figure}[h!]
\centering
\includegraphics[width=0.8\textwidth]{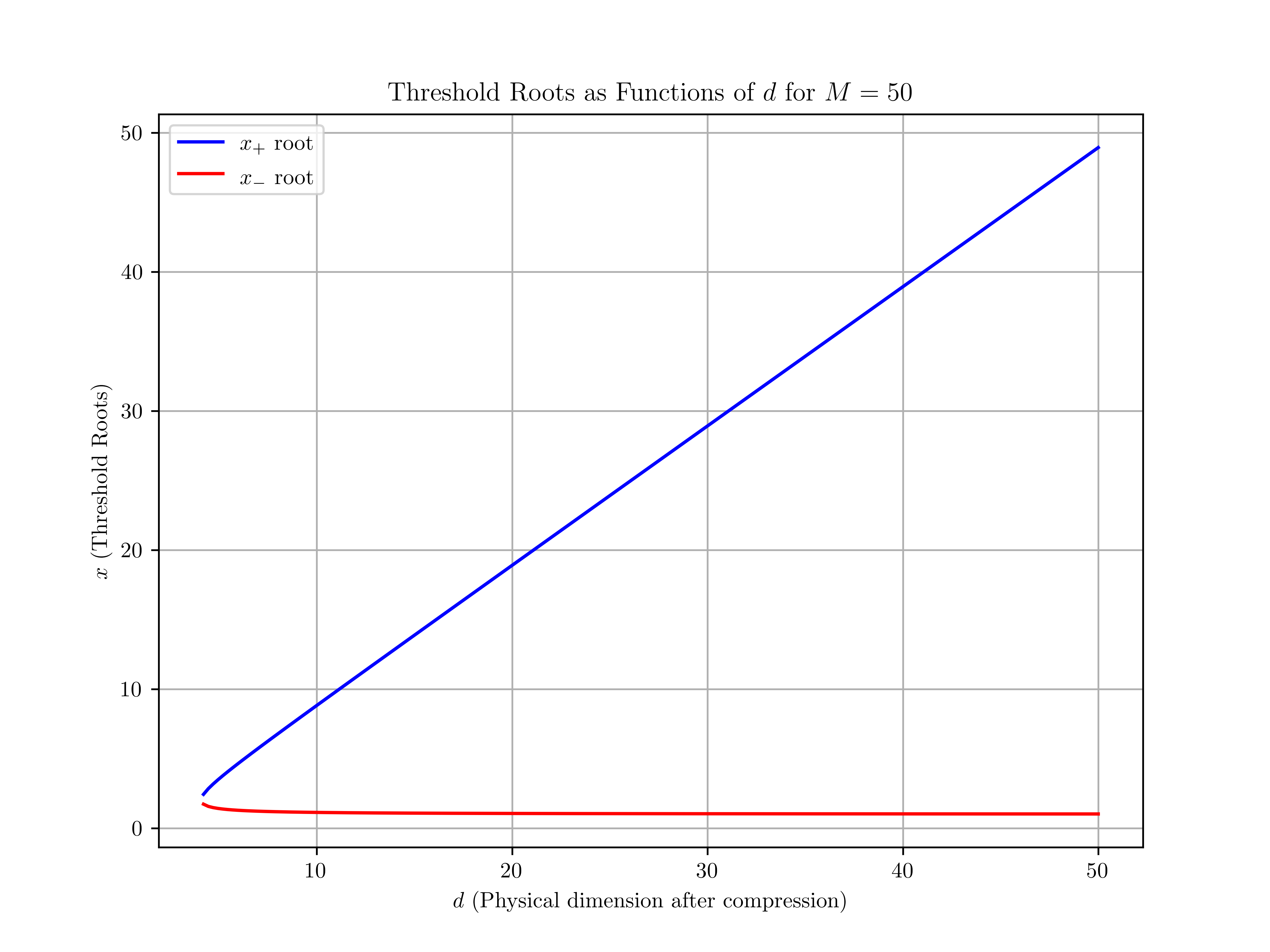}
\caption{Threshold roots \( x_+ \) and \( x_- \) as a function of \( d \) for \( M = 50 \). The curves depict the conditions under which a comb tensor network outperforms a regular MPS. As \( d \) grows, the beneficial regime for the comb structure becomes apparent, guiding the selection of parameters for efficient generative modeling of continuous data.}
\label{fig:output}
\end{figure}

The plot in Fig. \ref{fig:output} shows the threshold roots \( x_{\pm}(d) \) as functions of the compressed physical dimension \( d \) for a fixed \( M = 50 \).
When \( d \) and \( M \) vary, these roots determine when switching from a traditional MPS to a comb-shaped tensor network reduces contraction complexity. In other words, for \( x_+>x > x_- \) (the smaller positive root), the comb network outperforms the MPS, particularly in handling continuous, high-dimensional data scenarios.

For $N=5, d=30, D=100, x=10, M=50$:
\[
x_- \approx 1.04, \quad x_+ \approx 28.83.
\]
Since $10$ is greater than $1.04$ but less than $28.83$, the Comb Tensor Network reduces contraction complexity relative to the regular MPS in this parameter regime.

\section*{Conclusion}
By deriving explicit thresholds for the bond dimension $x$ in relation to the compressed dimension $d$, we have shown that the Comb Tensor Network can significantly lower contraction complexity compared to a traditional MPS. Since contraction is the primary computational step in training and evaluating tensor network models, improving its efficiency is paramount, especially when dealing with continuous, high-dimensional data.

Our analysis reveals two critical roots, $x_-$ and $x_+$, which define a regime where the Comb Tensor Network outperforms the conventional MPS. In particular, when $x$ lies between $x_-$ and $x_+$, the comb architecture leverages both compression and its distinctive geometry to achieve greater efficiency. This result provides a clear guideline for choosing parameters to ensure more efficient computations and learning tasks in challenging, large-scale scenarios.

The implementation of the compression layer discussed in this paper is available on \href{https://github.com/DanyloKolesnyk/QuantumMLChronicles}{our GitHub repository} \cite{github}. We will be posting more detailed plots that cover not only contraction complexities but also optimization processes in the near future. We encourage interested readers to visit the repository for updates and to contribute to the ongoing development of these tensor network models.

\bibliographystyle{unsrt}

\begin{thebibliography}{1}
\bibitem{meiburg2023generative}
A.~Meiburg, J.~Chen, J.~Miller, R.~Tihon, G.~Rabusseau, and A.~Perdomo-Ortiz,
\newblock Generative Learning of Continuous Data by Tensor Networks.
\newblock \href{https://arxiv.org/abs/2310.20498}{arXiv:2310.20498}.
\bibitem{white1992}
S.~R. White,
\newblock ``Density matrix formulation for quantum renormalization groups,''
\newblock {\em Physical Review Letters}, vol.~69, no.~19, pp. 2863--2866, 1992.
\newblock \href{https://doi.org/10.1103/PhysRevLett.69.2863}{doi:10.1103/PhysRevLett.69.2863}.

\bibitem{schollwoeck2005}
U.~Schollwöck,
\newblock ``The density-matrix renormalization group,''
\newblock {\em Reviews of Modern Physics}, vol.~77, no.~1, pp. 259--315, 2005.
\newblock \href{https://doi.org/10.1103/RevModPhys.77.259}{doi:10.1103/RevModPhys.77.259}.

\bibitem{schollwoeck2011}
U.~Schollwöck,
\newblock ``The density-matrix renormalization group in the age of matrix product states,''
\newblock {\em Annals of Physics}, vol.~326, no.~1, pp. 96--192, 2011.
\newblock \href{https://doi.org/10.1016/j.aop.2010.09.012}{doi:10.1016/j.aop.2010.09.012}.

\bibitem{chepiga2019}
N.~Chepiga and S.~R. White,
\newblock ``Comb tensor networks,''
\newblock {\em Physical Review B}, vol.~99, no.~23, p. 235426, 2019.
\newblock \href{https://doi.org/10.1103/PhysRevB.99.235426}{doi:10.1103/PhysRevB.99.235426}.

\bibitem{li2020}
Z.~Li,
\newblock ``Expressibility of comb tensor network states (CTNS) for the P-cluster and the FeMo-cofactor of nitrogenase,''
\newblock {\em arXiv preprint arXiv:2009.12573}, 2020.
\newblock \href{https://arxiv.org/abs/2009.12573}{arXiv:2009.12573}.

\bibitem{verstraete2008}
F.~Verstraete, V.~Murg, and J.~I. Cirac,
\newblock ``Matrix product states, projected entangled pair states, and variational renormalization group methods for quantum spin systems,''
\newblock {\em Advances in Physics}, vol.~57, no.~2, pp. 143--224, 2008.
\newblock \href{https://doi.org/10.1080/14789940801912366}{doi:10.1080/14789940801912366}.

\bibitem{github}
\href{https://github.com/DanyloKolesnyk/QuantumMLChronicles}{https://github.com/DanyloKolesnyk/QuantumMLChronicles}

\end{thebibliography}

\end{document}